\documentclass[runningheads]{llncs}
\usepackage{booktabs}      
\usepackage{multirow}      
\usepackage{makecell}      
\usepackage[table]{xcolor} 
\usepackage{colortbl}      
\usepackage{array}         
\usepackage{graphicx}
\usepackage{amsmath}
\usepackage{amssymb}   
\definecolor{bestcolor}{RGB}{244,190,194}
\definecolor{secondcolor}{RGB}{245,220,170}

\begin{document}
\title{Fast and Lightweight Novel View Synthesis with Differentiable Multiplane Image}

\titlerunning{Fast Lightweight NVS with MPI}
%
\author{Kaidi Zhang\inst{1} \and
Guanxu Zhu\inst{2}}


\institute{
Universiti Malaya \and Wuhan University \\
\email{kaidizhang0813@gmail.com,zhuguanxu@whu.edu.cn}\\
}
%
\maketitle

%

\begin{abstract}
Recently, novel view synthesis has witnessed remarkable progress, with mainstream methods such as Neural Radiance Fields (NeRF) and 3D Gaussian Splatting (3DGS) delivering impressive results.
However, these approaches often struggle to balance rendering speed and model size, and their optimization-based training can be highly time-consuming. Furthermore, they typically rely on dense observations, often failing to produce satisfactory results under sparse-view conditions. 
Although feed-forward reconstruction significantly reduces the optimization time of 3DGS, its pixel-aligned formulation generates millions of Gaussians from a single image, severely limiting its practical deployment on mobile devices.
To address these limitations, we revisit the Multiplane Image (MPI) representation, which represents scenes using a compact set of planar layers for efficient novel view synthesis.
Leveraging recent advances in visual foundation models, we utilize predicted point maps for reliable geometric initialization, followed by differentiable optimization. To address the issues of holes and artifacts in sparsely initialized MPI, we introduce one-step diffusion, which participates in both the differentiable optimization of MPI and the post-processing of rendering results.
Compared with a representative GS-based method, our approach is 30.7\% faster and uses only 14.8\% of its model size, while achieving competitive synthesis quality on front-view scenarios.

\end{abstract}

\section{Introduction}
\label{sec:introduction}
Sparse-view reconstruction significantly bridges the gap between 2D imagery and immersive 3D experiences, playing a crucial role in downstream applications such as robotics and 3D photography. The practical deployment of 3D representations imposes stringent requirements on both model compactness and rendering speed.

Recent advances, such as Neural Radiance Fields (NeRF~\cite{mildenhall2021nerf}) and their variants, have demonstrated impressive capabilities in synthesizing high-quality novel views. However, optimizing the continuous volumetric fields of NeRF typically relies on dense multi-view observations. As an alternative explicit representation, 3D Gaussian Splatting(3DGS~\cite{kerbl20233d}) has gained increasing attention due to its exceptional rendering efficiency. Despite this advantage, early 3DGS methods share a similar limitation with NeRF, requiring densely sampled viewpoints to maintain rendering fidelity. More recently, feed-forward paradigms have been introduced to directly infer 3D Gaussian representations in a single pass. While these approaches achieve notable progress, most of them adopt pixel-aligned Gaussian regression, where each pixel corresponds to one Gaussian. This design inevitably leads to severe redundancy— for instance, just a few sparse-view images can generate millions of Gaussians, rendering such methods impractical for deployment on resource-constrained devices like mobile platforms.

To address the limitations arising from high resource consumption and sparse input views, we turn our attention to Multiplane Images (MPI), which provide an efficient discrete representation for modeling continuous scenes. MPI-based approaches ~\cite{shade1998layered,szeliski1999stereo,zhou2018stereo} approximate a 3D scene using a set of fronto-parallel planes within a camera frustum, offering higher efficiency and lower computational cost compared to continuous representations like NeRF, albeit at the expense of representational continuity. 
However, existing MPI-based methods are typically restricted to the camera frustum and struggle to handle disocclusions and artifacts caused by viewpoint changes. This limitation largely stems from their reliance on feed-forward predictions (e.g., CNN-based models) from a single image, where performance is bounded by dataset scale and model capacity. Additionally, these methods are prone to depth discretization artifacts and repetitive texture patterns, which limit their applicability in real-world scenarios. Benefiting from recent advances in depth estimation and vision foundation models (e.g., VGGT~\cite{wang2025vggt}), it is now possible to infer geometric structure from few images. The resulting point clouds provide a strong geometric prior for scene representation, thus enabling the differentiable optimization of MPI.

Building upon these developments, we propose our Multi-view MPI, a method that leverages a learnable MPI representation. Compared to prior MPI-based approaches, our method significantly enhances 3D scene modeling capacity and supports a broader range of camera motions. Instead of hallucinating occluded regions through plane generation, which often yields low-quality texture completion, we treat MPI as a planar representation and explicitly optimize it using multi-view supervision to recover missing geometry. A key distinction of our approach lies in representing MPI as learnable parameters that are directly optimized via volume rendering, acknowledging that extended 3D scenes are too complex to be reliably inferred in a single forward pass. This formulation enables both flexible viewpoint control and efficient, 3D-consistent novel view synthesis, while effectively mitigating repetitive texture artifacts. 

To further address common MPI issues (e.g., artifacts), we leverage recent advances in generative models and integrate DIFIX~\cite{wu2025difix3d}, a fast one-step diffusion model, into our pipeline in two ways: First, we interpolate sparse-view rendering results, feed them into the diffusion model for restoration, and retrain the MPI therewith to distill the strong prior into it. Second, during rendering and inference, the diffusion model acts as a post-processor to enhance rendering quality. In summary, our main contributions are as follows:
\begin{itemize}
    \item We propose Multi-view MPI, a novel view synthesis framework that enables broader-range viewpoint exploration and fast, 3D-consistent rendering. Besides, we formulate MPI as a learnable planar representation that supports the differentiable optimization from multiple views.
    \item We incorporate a Neural Enhancer into both the optimization process and post-processing to address common MPI issues such as artifacts, thereby forming a more robust pipeline.
    \item We conduct extensive experiments on multiple benchmarks, including the LLFF real-world dataset and the NeRF synthetic dataset. Our method achieves the best PSNR among representative novel view synthesis methods while maintaining a compact representation and high rendering efficiency.
\end{itemize}

\section{Related Work}
\label{sec:related_work}
\subsection{Novel View Synthesis and 3D Representation. }
NeRF~\cite{mildenhall2021nerf} and its related works~\cite{gao2022nerf,chen2021mvsnerf,barron2021mipnerf} have achieved remarkable progress in NVS. To achieve
faster training and rendering, NeRFs have gradually incorporated explicit structures~\cite{chen2022tensorf,mueller2022instant}, while remaining computationally expensive due to the dense sampling along each ray. In
contrast, 3DGS~\cite{kerbl20233d} adopts a purely explicit representation, achieving photorealistic rendering quality while significantly improving rendering speed. Subsequent works have further enhanced 3DGS
by incorporating techniques such as anti-aliasing \cite{yu2024mipsplatting}, training acceleration, and block-wise
optimization~\cite{liu2024citygaussian,lin2024vastgaussian}. 
Recently, feed-forward Gaussian methods~\cite{charatan2024pixelsplat} have gradually replaced the traditional reconstruction paradigm. 
This paradigm has achieved remarkable results in sparse-view and single-view 3D reconstruction.
However, such feed-forward approaches are often pixel-aligned: each pixel is decoded into a Gaussian, which leads to severe redundancy among Gaussians. Although methods such as AnySplat~\cite{jiang2025anysplat} attempt to eliminate redundancy in multi-view pixel-aligned reconstruction using voxels, they still yield point clouds with hundreds of thousands of Gaussians. Consequently, the aforementioned 3D representations impose high requirements on computing devices.

\subsection{Layer-based Methods}
Layer-based methods represent a 3D scene using discrete layers, resulting in higher efficiency and lower computational cost compared to continuous 3D representations like NeRF. Layer-based methods,
such as 3D-photography~\cite{shih20203d} and SLIDE~\cite{jampani2021slide}, can produce high-quality synthesis results from a single
input image.
As a discrete diffuse volumetric scene representation, MPI-based
methods~\cite{shade1998layered,szeliski1999stereo,zhou2018stereo} were proposed to represent the scene using multiple planes in a frustum. 
Srinivasan et al.~\cite{srinivasan2019pushing} proposed a two-step MPI prediction procedure to alleviate issues of depth discretization artifacts and repeated texture artifacts that appear in MPIs rendering
results. NeX~\cite{wizadwongsa2021nex} extended MPIs for non-Lambertian material and was able to handle view-dependent effects
effectively. DeepView~\cite{flynn2019deepview} incorporated occlusion
reasoning to generate an MPI from a set of sparse camera viewpoints. 
However, these
methods heavily rely on large-scale datasets to train MPI predictors;
thus, the MPI prediction quality is highly constrained by the training data and can hardly adapt to new manifolds.

\subsection{Sparse view reconstruction}
To circumvent the requirement of dense input views, a line of NeRFs~\cite{chen2021mvsnerf,chibane2021srf,liu2022neuralrays,trevithick2021grf,wang2021ibrnet,yu2021pixelnerf} incorporates prior knowledge by pre-training conditional models of radiance fields on large posed multi-view datasets. Despite showing promising results on sparse input images, their generalization to out-of-distribution novel views remains a challenge. Sparse-view reconstruction for 3DGS is typically achieved in a feed-forward manner. In particular, pixelSplat~\cite{charatan2024pixelsplat} and Splatter Image~\cite{szymanowicz2024splatterimage} predict 3D Gaussians directly from image features, while MVSplat~\cite{chen2024mvsplat} encodes feature matching information using cost volumes and yields more accurate geometry. However, it inherently suffers from the limitations of feature matching in challenging scenarios such as texture-less regions and reflective surfaces. Besides, a series of MPI-based methods (SVMPI~\cite{tucker2020single}, VMPI~\cite{li2020synthesizing}, and AdaMPI~\cite{han2022single}) train convolutional image generators to infer occluded geometry in 3D space, but they do not exploit diffusion-based restoration priors for MPI optimization or inference, which limits their ability to handle artifacts and missing regions under sparse-view settings.

\section{Method}
\label{sec:method}
\vspace{-.7cm}
\begin{figure}
    \centering
    \includegraphics[width=1\linewidth]{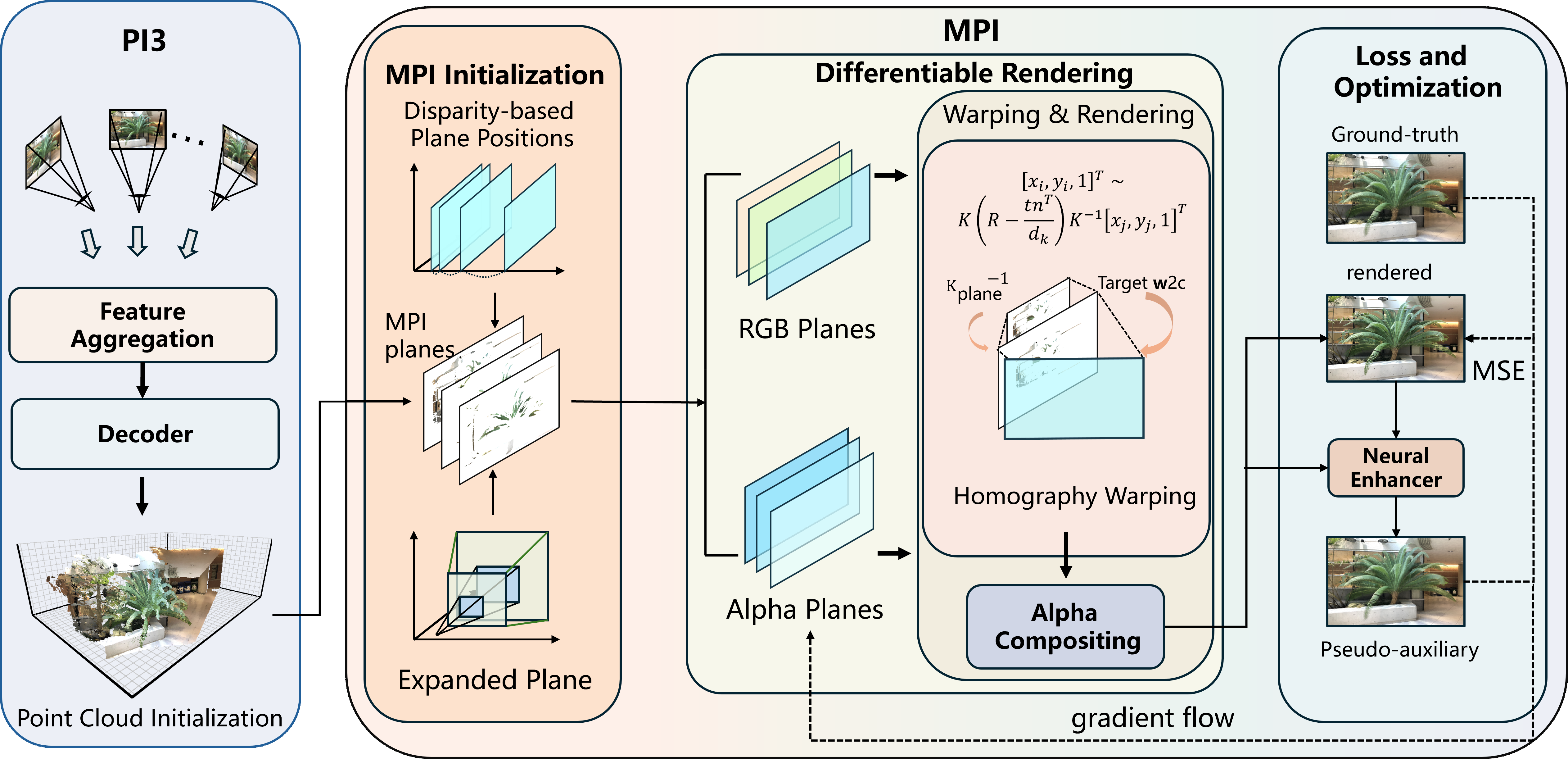}
    \caption{Overview of the proposed Multi-view MPI framework. Given sparse-view input images, PI3 first estimates camera poses, depth maps, and point clouds for reliable MPI initialization. The initialized expanded MPI is then optimized through differentiable rendering and multi-view supervision. To further improve rendering quality, the DIFIX-based neural enhancer is incorporated both during optimization for pseudo-view refinement and during inference as a post-processing module.}
    \label{fig:pipeline}
    \vspace{-.5cm}
\end{figure}

This section presents the full pipeline of our method. Sec.~\ref{sec:foundation_init} introduces the foundation-model-based geometric initialization, which provides reliable priors for MPI construction. Sec.~\ref{sec:expanded_mpi} then builds upon this initialization to construct and optimize an expanded learnable MPI through differentiable rendering. Sec.~\ref{sec:neural_enhancer} further refines the rendered views using a single-step diffusion neural enhancer, improving both optimization supervision and inference quality.

\subsection{Large Foundation Model for MPI Initialization.}
\label{sec:foundation_init}
Due to the lack of reliable geometric initialization in raw images, most previous methods generate planar representations using CNNs, an approach constrained by data volume and model scale. Benefiting from large reconstruction models such as VGGT~\cite{wang2025vggt} and PI3~\cite{wang2026pi3}, estimating reliable sparse-view geometry from few input images has become increasingly feasible.
PI3 employs a permutation-equivariant architecture to mitigate detrimental inductive biases. In contrast, prior models predominantly rely on a fixed reference view in visual geometry reconstruction, which enables PI3 to achieve superior reconstruction performance. 


Specifically, given an input image set \(\mathcal{I}=\{I_1,I_2,\dots,I_n\}\), PI3 estimates the poses \([\mathbf{p}_1,\mathbf{p}_2,\dots,\mathbf{p}_n]\) of these images. Concurrently, for each input image, PI3 outputs a dense depth map and per-pixel confidence scores.

Given a depth map \(D_i\), the 3D point corresponding to a pixel location \(\mathbf{u}=[u,v]^T\) can be obtained by back-projecting it into the camera coordinate system:
\begin{equation}
\mathbf{X}_i^c(\mathbf{u}) = D_i(\mathbf{u})\, K^{-1}\tilde{\mathbf{u}},
\quad
\tilde{\mathbf{u}} = [u,v,1]^T,
\label{eq:backprojection}
\end{equation}
where \(D_i(\mathbf{u})\) denotes the depth value at pixel \(\mathbf{u}\), \(K\) is the camera intrinsic matrix, and \(\mathbf{X}_i^c(\mathbf{u})=[X,Y,Z]^T\) is the corresponding 3D point in the camera coordinate system. Using the estimated camera poses, we transform all point clouds to the world coordinate system. After aligning all point clouds to the world coordinate system, we segment the original image based on depth and project it onto multiple planes distributed along the depth axis.

\subsection{Expanded MPI and Differentiable Rendering}
\label{sec:expanded_mpi}
Since the observation range of multi-view inputs is often larger than the reference range of a single-view frustum, it is necessary to expand the reference frustum to cover a larger spatial range. We propose to use larger planes for the view frustum with an expanded view angle \( \theta \), while rendering only the region corresponding to the camera's original field of view. The relationship between the expanded frustum angle and the plane size is defined as \( \frac{w \cdot a}{2f} = \tan\left( \frac{\theta}{2} \right) \), where \( w \) represents the length of the longer side of the original plane, \( a \) denotes the expansion factor, \( f \) is the camera focal length, and \( \theta \) is the expanded view angle. During testing, we perform homography transformation on the expanded MPI and apply volume rendering only to the parameters within the target camera's field of view, rather than rendering the entire expanded planes.

\subsubsection{Renderer and Optimization.}
\label{sec:renderer_optimization}
The original method using MPIs represents the 3D scene as a series of parallel planes in a view frustum. The plane homography transformation~\cite{Hartley2003} is defined by:
\begin{equation}
\left[x_{i}, y_{i}, 1\right]^{T} \sim K\left(R-\frac{t n^{T}}{d_{k}}\right) K^{-1}\left[x_{j}, y_{j}, 1\right]^{T},
\label{eq:mpi_homography}
\end{equation}
where $n = [0, 0, 1]^{T}$ is the plane normal, $d_{k}$ is the depth of the plane $k$, $x$ and $y$ denote the coordinates of a pixel in the target view $j$ or source view $i$, respectively, $K$ denotes the camera intrinsics, and $R$ and $t$ are the rotation and translation matrices, respectively.

The layers of our MPI are distributed evenly along the $z$-axis of the entire 3D space. We render the target view through volume rendering using the equations:
\begin{equation}
I = \sum_{i=1}^{N} T_{i}\left(1 - \exp\left(-\sigma_{d_{i}} \delta_{d_{i}}\right)\right) c_{d_{i}},
\label{eq:mpi_render1}
\end{equation}
\begin{equation}
T_i = \exp\left(-\sum_{j=1}^{i-1}\sigma_{d_j}\delta_{d_j}\right),
\label{eq:mpi_render2}
\end{equation}

Since the aforementioned rendering process is fully differentiable, we choose to construct MPI as learnable parameters instead of using an MPI predictor. Therefore, the MPI representation can be directly optimized from multi-view images via differentiable rendering and gradient descent. This novel approach allows for improved optimization efficiency and faster convergence rates since the rendering of rays can be efficiently parallelized.

\vspace{-.5cm}

\subsubsection{Discussion.}
Different from the training and rendering pipeline of the original NeRF, the MPI-based method operates in a discrete manner. It eliminates the need for sampling along rays and feeding them into MLP for intensive computation and volume rendering. Unlike NeRF, which requires MLP to decode features to derive density and color, MPI stores density and color directly within the planar pixel information and directly performs integration along rays. Consequently, the MPI representation bypasses the MLP decoding process, enabling direct integration and thus achieving extremely fast rendering speed.
\subsection{Neural Enhancer with Single-Step Diffusion Models}
\label{sec:neural_enhancer}
\begin{figure}[h]
    \vspace{-.5cm}
    \centering 
    \includegraphics[width=0.8\textwidth]{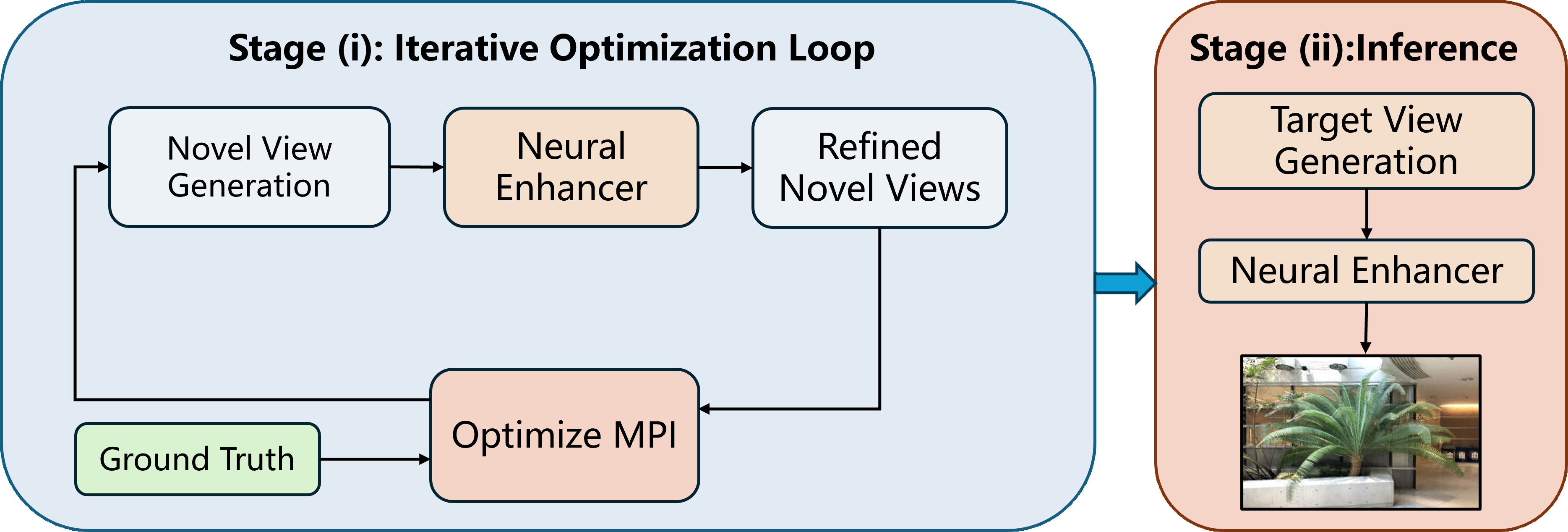} 
    \caption{ We adopt the Neural Enhancer to provide extra auxiliary views during optimization and serve as a lightweight backend processor in the rendering stage. Benefiting from its one-step inference property, our method can maintain a real-time rendering performance.} 
    \label{fig:difix} 
    \vspace{-.9cm}
\end{figure}
\vspace{-.1cm}
\subsubsection{Diffusion Models.} DMs~\cite{ho2020denoising,song2020improved,dhariwal2021diffusion} learn to model the data distribution $p_{\text{data}}(\mathbf{x})$ through iterative denoising and are trained with denoising score matching~\cite{ho2020denoising,vincent2008extracting,hyvarinen2005estimation,song2020improved,song2019generative,dhariwal2021diffusion,song2021score}. Specifically, to train a diffusion model, diffused versions $\mathbf{x}_\tau = \alpha_\tau \mathbf{x} + \sigma_\tau \boldsymbol{\epsilon}$ of the data $\mathbf{x} \sim p_{\text{data}}$ are generated, by progressively adding Gaussian noise $\boldsymbol{\epsilon} \sim \mathcal{N}(\mathbf{0}, \mathbf{I})$. Learnable parameters $\theta$ of the denoiser model $\mathbf{F}_\theta$ are optimized using the denoising score matching objective:
\begin{equation}
\mathbb{E}_{\mathbf{x} \sim p_{\text{data}}, \tau \sim p_\tau, \boldsymbol{\epsilon} \sim \mathcal{N}(\mathbf{0}, \mathbf{I})} \left[ \left\| \mathbf{y} - \mathbf{F}_\theta(\mathbf{x}_\tau; \mathbf{c}, \tau) \right\|_2^2 \right],
\label{eq:diffusion_objective}
\end{equation}

where $\mathbf{c}$ represents optional conditioning information, such as a text prompt or image context. Depending on the model formulation, the target vector $\mathbf{y}$ is usually set as the added noise $\boldsymbol{\epsilon}$. Finally, $p_\tau$ denotes a uniform distribution over the diffusion time variable $\tau$. 

For the novel view synthesis task, DIFIX~\cite{wu2025difix3d} is a single-step image diffusion model trained to enhance and remove artifacts in rendered novel views caused by under-constrained regions of the 3D representation. It is trained on a large dataset consisting of pairs of artifact-corrupted images (containing typical artifacts in novel-view synthesis) and their corresponding clean ground-truth images. Therefore, it is effective for improving rendering quality.
\vspace{-.1cm}
\subsubsection{Diffusion Models as Neural Enhancer.} 
Due to the compactness of the model representation, the quality of synthesized views is typically high for small viewpoint changes. However, when the viewpoint variation becomes large, the pseudo views generated by our method tend to be overly sharp and prone to noise points as well as flickering artifacts.

To mitigate these artifacts, we leverage the strong generative prior of DIFIX, which plays two roles in our pipeline:

(i) Optimization: Iteratively augmenting the training set with clean pseudo-views to improve the underlying 3D representation in unobserved areas. Given an intermediate MPI representation, we render novel views and feed them to DIFIX, which acts as a neural enhancer to remove artifacts and improve the quality of noisy rendered views. The camera poses selected for rendering novel views can be obtained via pose interpolation, gradually approaching the target poses from the reference ones.

(ii) Inference: A real-time post-processing step that further reduces artifacts caused by insufficient or inconsistent training supervision.
\vspace{-.1cm}
\subsubsection{Overall training objective.}

We optimize the parameters by minimizing the MSE loss between the rendered images and their corresponding supervision images. Since the entire optimization process involves supervision from both the original ground-truth images and the refined novel-view images, it is necessary to balance the weights of different types of supervision to ensure consistency during practical optimization. 
\begin{equation}
\mathcal{L} = \mathcal{L}_{\mathrm{gt}} + \alpha \mathcal{L}_{\mathrm{pseudo}},
\label{eq:overall_loss}
\end{equation}
where \(\mathcal{L}_{\mathrm{gt}}\) denotes the MSE loss between the rendered views and the original ground-truth images, and \(\mathcal{L}_{\mathrm{pseudo}}\) denotes the MSE loss between the rendered pseudo views and their refined counterparts produced by the neural enhancer. In our practice, we set \(\alpha = 0.8\) to further encourage geometric consistency while maintaining fidelity to the original observations.
\vspace{-.2cm}
\section{Experiments}
\label{sec:experiments}

\begin{figure}[!htbp]
\vspace{-.7cm}
    \centering
    \includegraphics[width=0.72\linewidth,
        height=0.42\textheight,
        keepaspectratio]{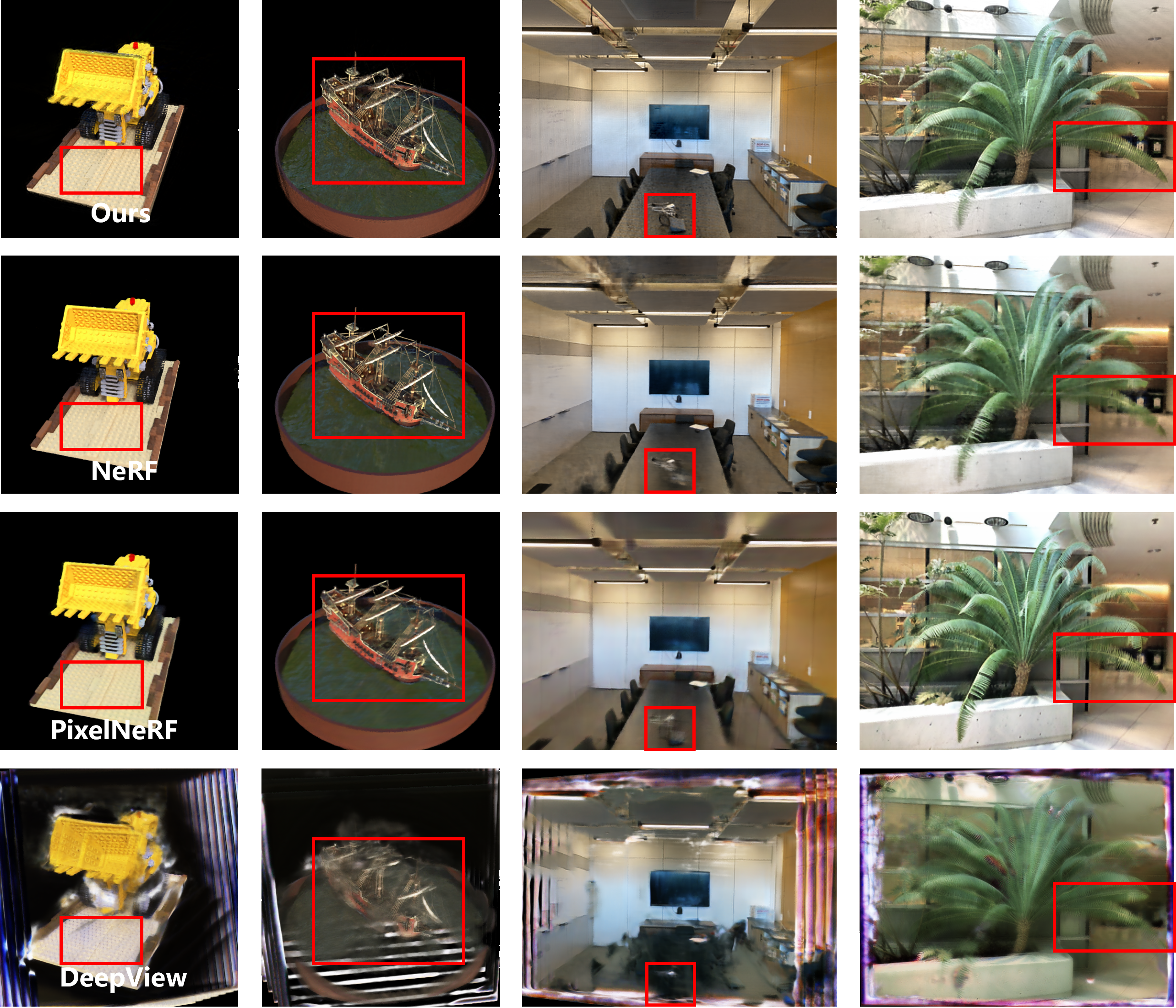}
    \caption{Qualitative comparison on representative novel-view synthesis scenes. Red boxes highlight challenging regions where our method produces sharper structures and fewer artifacts than NeRF, PixelNeRF, and DeepView.}
    \label{fig:qualitative_comparison}
    \vspace{-.4cm}
\end{figure}

To ensure a consistent evaluation, all methods are evaluated using input images at the same resolution. For methods that require geometric initialization, we use the same pre-trained multi-view model when applicable. We initialize our scene representation with the point cloud estimated by PI3~\cite{wang2026pi3}, and align it with the ground-truth poses via SIM(3).


\subsection{Experimental Setup}
\subsubsection{Datasets and Experimental Settings.}
We compare our method with representative novel view synthesis methods, including NeRF~\cite{mildenhall2021nerf}, PixelNeRF~\cite{yu2021pixelnerf}, DeepView~\cite{flynn2019deepview}, and AnySplat~\cite{jiang2025anysplat}. NeRF is a per-scene optimization method based on neural radiance fields, while PixelNeRF is a generalized sparse-view model that conditions radiance field prediction on input images. DeepView is included as a representative MPI-based baseline that predicts layered scene representations from sparse input views. AnySplat is a recent feed-forward 3DGS method for sparse-view reconstruction. Unlike feed-forward methods that directly predict scene representations in a single pass, our method optimizes a learnable MPI representation for each scene using multi-view supervision.
\vspace{-.2cm}
\subsubsection{Metrics.} 
We assess performance using several commonly used image quality evaluation metrics, including SSIM, PSNR, and LPIPS~\cite{zhang2018lpips}.
\vspace{-.2cm}
\subsubsection{Implementation Details.}
All experiments are implemented in PyTorch~\cite{paszke2017pytorch} with the Adam~\cite{kingma2014adam} optimizer. We train each scene for 8 epochs with 32 MPI planes, using a learning rate of $10^{-3}$. For PI3 initialization, we retain predicted points with confidence scores greater than 0.5. During training, we apply DIFIX refinement once after each epoch to refine rendered pseudo-views and provide additional supervision for MPI optimization. 

For a fair comparison, DIFIX is disabled as an inference-time post-processing module for the quantitative results in Table~\ref{tab:table_methods}, where the MPI renderer runs at 135.62 FPS. When optionally enabled, DIFIX further enhances visual quality, while the full pipeline still runs at around 15 FPS on average.
\begin{figure}[!htbp]
    \centering
    \includegraphics[width=0.8\linewidth]{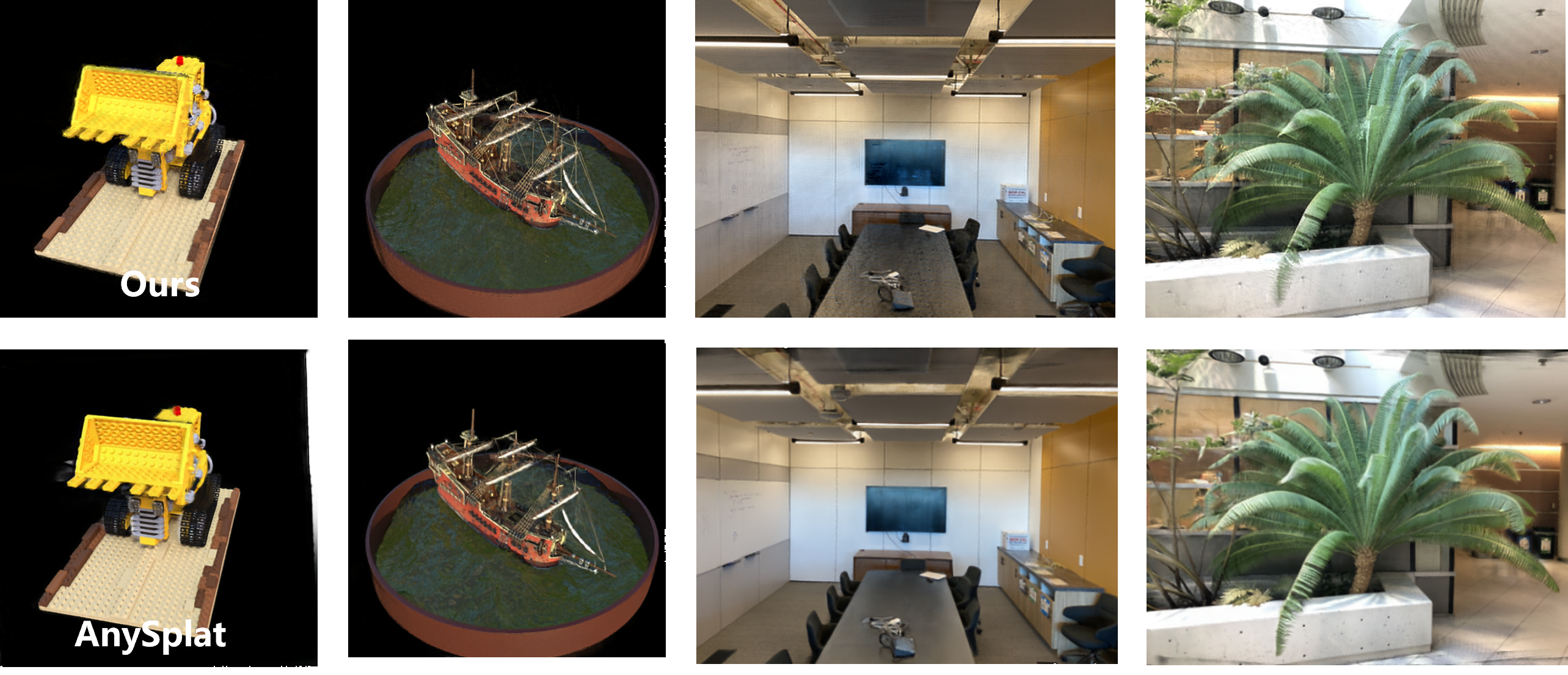}
    \caption{Compared with AnySplat, our method uses a more compact representation and achieves faster rendering, while maintaining competitive rendering quality.}
    \label{fig:mpi_comparison_AnySplat}
    \vspace{-.6cm}
\end{figure}

\vspace{-.3cm}
\subsection{Comparison in Novel View Synthesis}
\begin{table}[t]
\caption{\textbf{Quantitative Comparison on LLFF and NeRF Synthetic.}
We compare different neural rendering and view synthesis methods in terms of efficiency, storage cost, and rendering quality. 
Higher FPS, higher PSNR/SSIM, lower LPIPS, and smaller model size indicate better performance. 
The best and second-best results are highlighted as \colorbox{bestcolor}{best} and \colorbox{secondcolor}{second-best}.}
\label{tab:table_methods}
\centering
\small
\setlength{\tabcolsep}{3.2pt}
\renewcommand{\arraystretch}{1.08}
\resizebox{\linewidth}{!}{
\begin{tabular}{l|cc|ccc|ccc}
\hline
\multirow{2}{*}{Method}
& \multicolumn{2}{c|}{Rendering / Storage}
& \multicolumn{3}{c|}{LLFF}
& \multicolumn{3}{c}{NeRF Synthetic} \\
& Speed$\uparrow$ 
& Size$\downarrow$
& PSNR$\uparrow$ 
& SSIM$\uparrow$ 
& LPIPS$\downarrow$
& PSNR$\uparrow$ 
& SSIM$\uparrow$ 
& LPIPS$\downarrow$ \\
\hline
Ours      
& \cellcolor{secondcolor}135.62FPS
& 22.83MB  
& \cellcolor{bestcolor}25.469 
& \cellcolor{secondcolor}0.8005 
& \cellcolor{secondcolor}0.2973 
& \cellcolor{bestcolor}29.161 
& \cellcolor{bestcolor}0.9266 
& 0.1210 \\

NeRF      
& 0.212FPS
& \cellcolor{bestcolor}14.35MB  
& \cellcolor{secondcolor}25.048  
& 0.7418 
& 0.3218 
& \cellcolor{secondcolor}27.854 
& \cellcolor{secondcolor}0.9163 
& \cellcolor{secondcolor}0.1119 \\

PixelNeRF 
& 0.0255FPS  
& 112.82MB 
& 23.729 
& 0.7067 
& 0.3394 
& 23.360 
& 0.8647 
& 0.1578 \\

DeepView  
& \cellcolor{bestcolor}433.04FPS 
& \cellcolor{secondcolor}16.39MB  
& 12.204 
& 0.3685  
& 0.5396 
& 11.252
& 0.2345 
& 0.5673 \\

AnySplat  
& 103.72FPS
& 153.85MB 
& 23.782
& \cellcolor{bestcolor}0.8049 
& \cellcolor{bestcolor}0.1657 
& 26.259 
& 0.8561 
& \cellcolor{bestcolor}0.1079 \\
\hline
\end{tabular}
}
\vspace{-.5cm}
\end{table}

\subsubsection{Evaluation on the LLFF Dataset.}
The LLFF dataset consists of forward-facing real-world scenes. For each scene, we select one image out of every eight frames as part of the test set, while the remaining images are subsampled by taking one out of every three frames to form the training set, thereby constructing a sparse-view input setting.

Benefiting from the efficiency of layer-based representations, which preserve image resolution and fine texture details, our method produces sharper and more detailed renderings than PixelNeRF and DeepView. As shown in Table~\ref{tab:table_methods}, our method achieves the best PSNR on both LLFF and NeRF Synthetic among the evaluated methods, while maintaining competitive SSIM and LPIPS.

Compared with AnySplat, our method improves the rendering speed from 103.72 FPS to 135.62 FPS and reduces the model size from 153.85 MB to 22.83 MB. Although AnySplat obtains better SSIM or LPIPS on some metrics, its feed forward 3DGS representation is substantially larger. In contrast, our method provides a more compact and faster representation while preserving strong rendering quality.


\definecolor{bestcolor}{RGB}{255,150,160}
\definecolor{secondcolor}{RGB}{255,205,140}

\vspace{-0.37cm}
\subsubsection{Comparison in NeRF Synthetic Dataset.}
Since MPI-based representations are mainly suited to sparse-view inputs and forward-facing camera motions, we construct a locally forward-facing evaluation setting on NeRF Synthetic. For each scene, we select one reference image and its adjacent 30 frames. Among them, 4 frames are used for testing, and half of the remaining frames are uniformly sampled to construct the sparse-view training set.

Fig.~\ref{fig:qualitative_comparison} provides qualitative comparisons with NeRF, PixelNeRF, and DeepView on representative scenes. PixelNeRF tends to produce noisy results due to MLP underfitting, while DeepView suffers from depth discretization artifacts and repeated texture artifacts caused by MPI prediction. In contrast, our method generates novel views with sharper structures and more coherent geometry. Table~\ref{tab:table_methods} shows that our method achieves the highest PSNR on NeRF Synthetic while using a much smaller model than AnySplat~\cite{jiang2025anysplat}.


\subsection{Ablation Study}
\begin{table}[!b]
\centering
\caption{Ablation study of the main components in our framework, evaluated on the LLFF dataset. Removing PI3 initialization, differentiable MPI optimization, or diffusion enhancement consistently degrades reconstruction quality, demonstrating the effectiveness of each design.}
\label{tab:ablation}
\setlength{\tabcolsep}{8pt}
\renewcommand{\arraystretch}{1.15}
\begin{tabular}{lcccc}
\toprule
\textbf{Setting} & \textbf{PSNR$\uparrow$} & \textbf{SSIM$\uparrow$} & \textbf{LPIPS$\downarrow$} & \textbf{PSNR Drop$\downarrow$} \\
\midrule
Full model        & \textbf{25.469} & \textbf{0.8005} & \textbf{0.2973} & -- \\
w/o PI3 init.     & 17.363 & 0.5338 & 0.5137 & 8.106 \\
w/o MPI optim.    & 15.040 & 0.4382 & 0.4741 & 10.429 \\
w/o diffusion     & 21.632 & 0.6211 & 0.3899 & 3.837 \\
\bottomrule
\end{tabular}
\vspace{-.5cm}
\end{table}
\begin{figure}[t]
    \centering
    \includegraphics[width=1\linewidth]{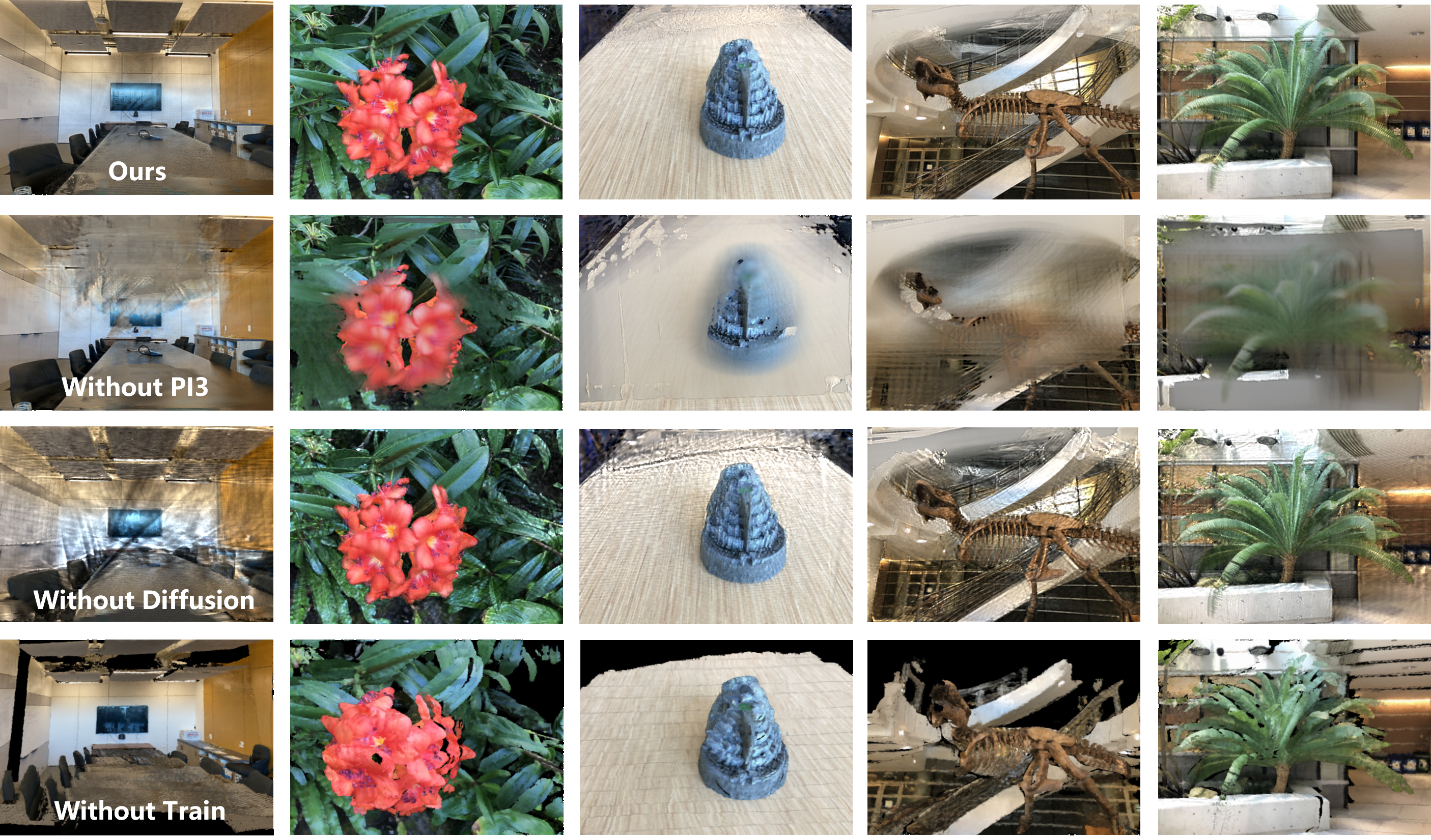}
    \caption{Visual ablation study on key components of the proposed method. Removing PI3 initialization, differentiable training, or diffusion enhancement leads to noticeable degradation in geometry reconstruction and image quality, including blur, structural distortion, and loss of fine details.}
    \label{fig:ablation}
\end{figure}
\subsubsection{MPI Model.}
 We remove the PI3 point cloud initialization and instead initialize all MPI planes from scratch to verify the effectiveness of our initialization strategy. In addition, we ablate the differentiable rendering optimization to validate the efficacy of our gradient-based optimization.

\subsubsection{Neural Enhancer.}
The Neural Enhancer serves two roles: on one hand, it acts as a powerful prior to assist the MPI optimization, addressing artifacts and missing regions; on the other hand, it works as a post-processing module to perform single-step denoising on the rendered novel-view images. Therefore, the ablation is conducted in two settings: (1) using the Neural Enhancer only as a post-processing module, and (2) employing the Neural Enhancer only for MPI optimization.
\vspace{-.3cm}
\section{Conclusion}
In this paper, we propose Multi-view MPI to address the high resource consumption associated with explicit representations and the rendering artifacts inherent in traditional MPI methods under wide camera motions. By formulating MPI as a learnable planar representation optimized via multi-view supervision, our approach significantly expands scene modeling capacity while effectively mitigating texture repetition. Furthermore, we integrate a one-step diffusion model as a Neural Enhancer to distill strong generative priors during optimization and refine outputs during inference. Consequently, our framework achieves high-quality, efficient, and 3D-consistent novel view synthesis.

\clearpage
\bibliographystyle{splncs04}
\bibliography{mybib}
\end{document}